%% file: template.tex
\documentclass{vgtc}                          

\graphicspath{{figures/}{pictures/}{images/}{./}} 

\usepackage{times}                     

\usepackage{tabularx}    
\usepackage{array}
\usepackage{booktabs}                  
\usepackage{lipsum}                    
\usepackage{mwe}                       

\usepackage{mathptmx}  
\PassOptionsToPackage{dvipsnames,svgnames}{xcolor}
\usepackage[most]{tcolorbox}

\onlineid{0}

\vgtccategory{Research}

\vgtcinsertpkg

\title{Multi-Agent Data Visualization and Narrative Generation}

\author{
Anton Wolter\thanks{e-mail:wol@cs.au.dk}\\ %
     \parbox{1.4in}  {\centering Aarhus University}\\ 
     \and
 Georgios Vidalakis \thanks{e-mail: gvid@kth.se}\\ %
     KTH Royal Institute of Technology \\ %
    \and 
    Michael Yu \thanks{e-mail: mjyu@kth.se}\\ %
      KTH Royal Institute of Technology \\ %
    \and 
     Ankit Grover\thanks{e-mail: agrover@kth.se}\\ %
      KTH Royal Institute of Technology \\ %
    \and 
    Vaishali Dhanoa \thanks{e-mail: dhanoa@cs.au.dk}\\ %
     \parbox{1.4in}{ \centering Aarhus University\\ %
    TU Wien}}
\teaser{
  \centering
  \includegraphics[width=\linewidth]{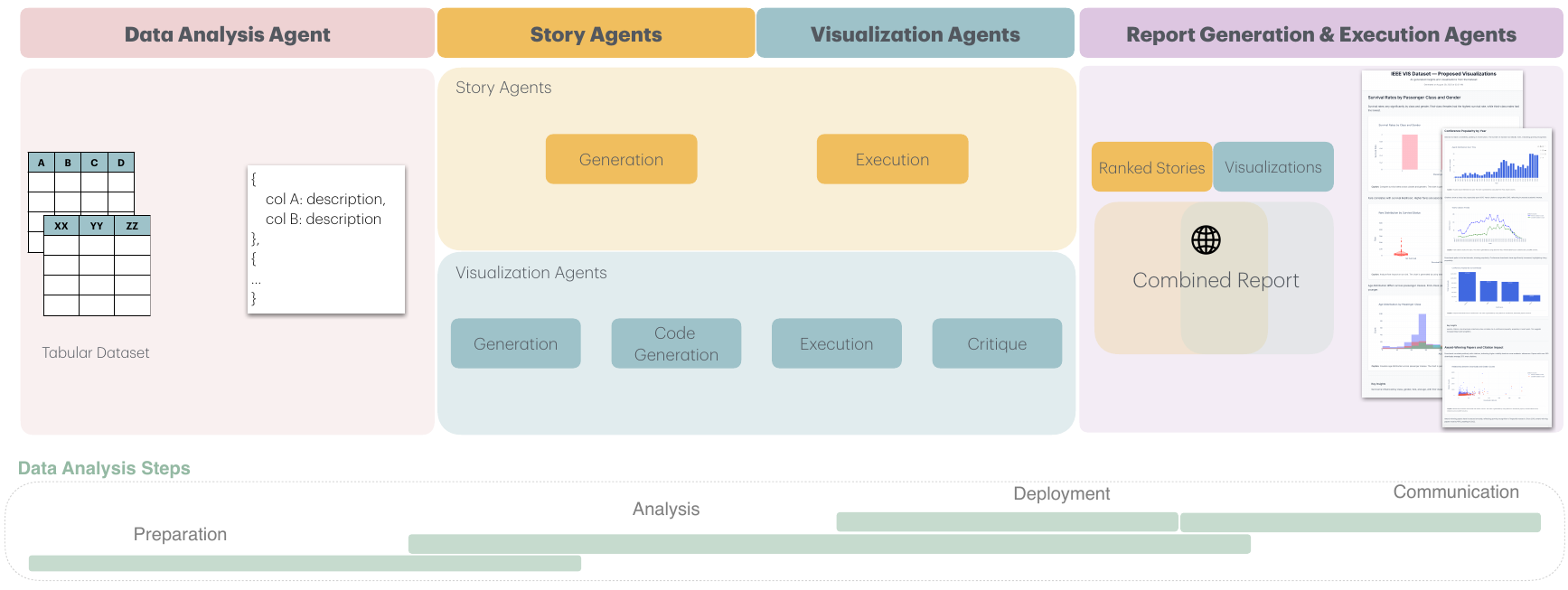} 
  \caption{\textbf{Overview of agents in our workflow}: Agent roles involved in our system throughout the data analysis workflow, based on Crisan et al.~\cite{crisan2021}}
  \label{fig:teaser}
}

\abstract{%
    \input{content/abstract}
} 


\begin{document}

\firstsection{Introduction \& Background}

\maketitle

\input{content/00-agent-roles-table}

\input{content/01-introduction}

\input{content/02-technical-concept}

\input{content/03-discussion}


\bibliographystyle{abbrv-doi}
\bibliography{template}

\end{document}

%% file: content/abstract.tex
Recent advancements in the field of AI agents have impacted the way we work, enabling greater automation and collaboration between humans and agents.
In the data visualization field, multi-agent systems can be useful for employing agents throughout the entire data-to-communication pipeline.
We present a lightweight multi-agent system that automates the data analysis workflow, from data exploration to generating coherent visual narratives for insight communication.
Our approach combines a hybrid multi-agent architecture with deterministic components, strategically externalizing critical logic from LLMs to improve transparency and reliability.
The system delivers granular, modular outputs that enable surgical modifications without full regeneration, supporting sustainable human-AI collaboration.
We evaluated our system across 4 diverse datasets, demonstrating strong generalizability, narrative quality, and computational efficiency with minimal dependencies.

%% file: content/00-agent-roles-table.tex
\definecolor{mypink}{HTML}{EED2D3}
\definecolor{myyellow}{HTML}{F0BE5E}
\definecolor{myblue}{HTML}{9EC2C9}
\definecolor{mypurple}{HTML}{DDC6E1}
\definecolor{mygreen}{HTML}{e5f5e0}

\newtcbox{\pinkbg}{on line,
  colback=mypink,
  colframe=mypink,
  arc=6pt,
  boxrule=0pt,
  left=2pt, right=2pt, top=2pt, bottom=2pt}

\newtcbox{\yellowbg}{on line,
  colback=myyellow,
  colframe=myyellow,
  arc=6pt,
  boxrule=0pt,
  left=2pt, right=2pt, top=2pt, bottom=2pt}

\newtcbox{\bluebg}{on line,
  colback=myblue,
  colframe=myblue,
  arc=6pt,
  boxrule=0pt,
  left=2pt, right=2pt, top=2pt, bottom=2pt}

\newtcbox{\purplebg}{on line,
  colback=mypurple,
  colframe=mypurple,
  arc=6pt,
  boxrule=0pt,
  left=2pt, right=2pt, top=2pt, bottom=2pt}

  \newtcbox{\greenbg}{on line,
  colback=mygreen,
  colframe=mygreen,
  arc=6pt,
  boxrule=0pt,
  left=2pt, right=2pt, top=2pt, bottom=2pt}

\begin{table*}[ht]
    \centering
    \small
    \caption{\textbf{Agent roles.}
        Our lightweight solution consists of several agents with different tasks that collaborate.}    
    \label{tab:agent_roles}
    \renewcommand{\arraystretch}{1.0}
    \begin{tabular}{p{4cm}p{7cm}p{5cm}}
        \toprule
        \textbf{Agent} & \textbf{Description} & \textbf{Input $\rightarrow$ Output} \\
        \toprule
        \pinkbg{Data Analysis Agent} & Analyzes tabular data to create structured metadata (domain-agnostic). &  CSV data source $\rightarrow$ JSON schema with metadata \\
        \midrule
        \yellowbg{Story Generation Agent} & Creates narrative ideas to support analysis goals. & JSON schema $\rightarrow$ Story ideas \\
        \midrule
        \yellowbg{Story Execution Agent} & Ranks narratives and integrates visualizations with text. & Story ideas and visualizations $\rightarrow$ Curated stories \\
        \midrule
        \bluebg{Visualization Generation Agent} & Proposes meaningful visualizations for the dataset. & JSON schema and story context $\rightarrow$ Vis ideas \\
        \midrule
        \bluebg{Code Generation Agent} & Transforms visualization ideas into executable code. & Visualization ideas $\rightarrow$ Plotly code \\
        \midrule
        \bluebg{Visualization Execution Agent} & Executes code to render charts. & Plotly code $\rightarrow$ Visualizations \\
        \midrule
        \bluebg{Visualization Critique Agent} & Evaluates charts against design principles and handles errors. & Generated visualizations $\rightarrow$ Refined visualizations \\
        \midrule
        \purplebg{Report Generation Agent} & Selects and orders content for final output. & Curated stories $\rightarrow$ Ordered content \\
        \midrule
        \purplebg{Report Execution Agent} & Renders the final presentation format. & Selected content $\rightarrow$ HTML report \\
        \midrule
        \greenbg{Monitoring Agent} & Tracks system performance and resource usage. & System events $\rightarrow$ Performance metrics \\
 
        \bottomrule
    \end{tabular}
\end{table*}

%% file: content/01-introduction.tex
Data-driven reports are routinely used across a wide range of domains, from large companies to small organizations where they are deployed for a large audience or catered to individual professionals. 
Creating such reports requires careful planning, in terms of data analysis and thoughtful design to deliver reports that effectively capture the sentiment of the data and presents compelling results in a meaningful way. 
They must also be continuously updated and maintained as new domains and data emerge. 
Despite the promises of commercial self-service BI tools that claim to offer a fast path to building data-driven reports, much of the effort still lies in preparing and transforming data, followed by the additional work of designing effective visualizations~\cite{walchshofer2023transitioning}.

With the rapid rise of AI agents, human-AI collaboration is getting reshaped across many domains, from co-creating or autonomously generating software code, creative writing, to even art production. 
These agents augment and amplify human capabilities~\cite{shneiderman2022human} by offloading manually intensive tasks from humans and enabling them to focus on critical decision making points. 
In the field of data visualization, the concept of Agentic Visualization, proposed by Dhanoa et al~\cite{dhanoa2025agentic}, emphasizes the role of various agents employed the data to communication pipeline.

We draw inspiration from several such works~\cite{wang2024dataformulator, zhao2024lightva} to present our lightweight solution that generates web-based data-driven reports with the help of multiple agents\footnote{Our workshop challenge submission report is available \href{https://purple-glacier-014f19d1e.6.azurestaticapps.net/api/output/1a2cdc37-4b8b-42f9-9403-fb0e66a63456}{on the submission server}.}.
In contrast to existing approaches~\cite{maddigan2023chat2vis, dibia2023lida, Narechania_2021} that use LLMs or other AI agents to create visualizations from data, we employ independent parallel agent processing with aggregation. Our compact approach automates the analysis of datasets and generates comprehensive data-driven reports that seamlessly integrate visualization and narrative explanations, as shown in Figure~\ref{fig:teaser}.We evaluate our approach for its generalizability across different datasets.

%% file: content/02-technical-concept.tex
\section{Technical Concept}

The architecture follows a role-based design pattern, as described by Dhanoa et al.~\cite{dhanoa2025agentic} where specialized agents handle distinct aspects of the visualization pipeline, from data preparation to visualization generation along with a narrative for the final report curation.
At the same time we strive for a compact and lightweight solution.~\cite{parnas1979extension,stevens1974structured}
Therefore solution leverages a custom Python-based node architecture with multiprocessing to orchestrate workflows, enabling automated visual report generation with data-driven narratives from tabular datasets with minimal third party technical dependencies.

Our system adopts a hybrid architecture that strategically externalizes critical logic from LLMs to deterministic components, improving transparency and reliability.
This composable design allows specialized modules to handle well-defined tasks while LLMs focus on high-level reasoning and narrative generation.
For creating data-driven reports, it uses several agents that either use LLMs or deterministic heuristics to produce relevant output.
We describe these agents and their tasks in table~\ref{tab:agent_roles}.
 
Plotly~\cite{plotly2015} serves as the primary charting library, supporting interactive visualizations that include line charts, bar plots, scatter plots, and heatmaps. 
The code generation pipeline creates Python scripts executed with error handling and retry mechanisms to ensure robustness.
For structured agentic inputs and outputs, Pydantic~\cite{pydantic} models are utilized.
Jinja2~\cite{jinja2} templating generates HTML reports that combine narrative explanations with embedded visualizations. 
The template structure includes modular components for story sections, chart integration, and an optional metrics display using responsive design principles.
The system enables monitoring of key metrics, such as story retention rate or LLM API calls with token usage, in order to provide transparency into the system's performance.
 
The implementation is generalized to analyze and generate reports for any tabular dataset. 
The dataset specification generation can be utilized for complementing different agent-based data analysis and communication workflows.
In our prompts we do not provide few-shot examples, and only rely on dataset-agnostic instructions.
Moreover, the prompts of the Visualization and Story agents can be easily adapted to different user groups. 
Finally, our implemented metrics collector can be used to collect granular feedback and analysis of performance for different datasets.
Evaluation across 4 diverse datasets (418-4,111 rows) achieved 75\% story retention for most datasets and 6-11 seconds per story runtime, validating generalizability and effectiveness.
Each dataset yielded proper reports with relevant visualizations and coherent narratives, demonstrating the system's domain-agnostic capabilities.

%% file: content/03-discussion.tex
\section{Discussion}

Our role-based multi-agent approach demonstrates promising capabilities for automated visualization generation, but some limitations warrant discussion.

\textbf{Strengths:} 
The multi-agent structure enables specialized processing at each workflow stage, improving output quality through focused expertise.
The orchestration layer ensures reliable state management and supports effective error recovery.
The modular architecture facilitates component-wise optimization and testing.
Notably, our system is lightweight: it relies on only one LLM dependency while being otherwise completely self-contained using standard popular dependencies from the Python ecosystem.
This design choice makes it comparatively easy to port into other infrastructures without other changes.

\textbf{Technical Limitations:} 
The system relies heavily on LLM consistency for code generation, with occasional syntax errors requiring retry mechanisms.
Missing semantic descriptions of the dataset columns can significantly affect the generated story ideas and the visualization interpretation.

\textbf{Future Directions:} 
While agents have demonstrated their capabilities in generating an automatic data-driven report, we believe that human-in-the-loop is required to validate and verify each intermediate step and ensure the validity of the results, especially in case of complex domain-specific datasets.
Additionally, the report could benefit from interactive capabilities to enable reader-driven discovery through exploration much beyond the AI's narrative.
Further exploration and research is also required to manage the resources effectively. To this end, we have incorporated a quantitative metrics for evaluation the failure rate of visualization generation, but it could be enhanced. We also believe that human intervention and feedback can significantly lower the error rates.